\documentclass[arts,article,submit,moreauthors,pdftex,10pt,a4paper]{mdpi2} 
\firstpage{1} 
\articlenumber{x}
\doinum{10.3390/------}
\pubvolume{xx}
\pubyear{2018}
\copyrightyear{2018}
\history{Received: date; Accepted: date; Published: date}

\Title{Expressivity in Natural and Artificial Systems}

\Author{Amy LaViers $^{1}$\orcidA{}*} 

\AuthorNames{Amy LaViers}  

\address{%
$^{1}$ \quad Mechanical Science and Engineering Department, University of Illinois at Urbana-Champaign}

\corres{Correspondence: alaviers@illinois.edu; +1 (217) 300-1486}

\abstract{ 
Roboticists are trying to replicate animal behavior in artificial systems. Yet, quantitative bounds on capacity of a moving platform (natural or artificial) to express information in the environment are not known. This paper presents a measure for the capacity of motion complexity -- the expressivity -- of articulated platforms (both natural and artificial) and shows that this measure is stagnant and unexpectedly limited in extant robotic systems.  This analysis indicates trends in increasing capacity in both internal and external complexity for natural systems while artificial, robotic systems have increased significantly in the capacity of computational (internal) states but remained more or less constant in mechanical (external) state capacity. This work presents a way to analyze trends in animal behavior and shows that robots are not capable of the same multi-faceted behavior in rich, dynamic environments as natural systems. 
}

\begin{document}

Abstractions about design goals such as form versus function help guide our thinking on constraining design spaces.  In robotics, optimization is often used to define a notion of success motion; yet, robots suffer from brittle behaviors that break down in dynamic environments.  In biology, natural selection is described via animal functionality inside a particular environmental context; yet, when individual animals are subjected to entirely new environments, they adapt \cite{turner2013biology}.  Consider pavement ants (\textit{Tetramorium caespitum}) brought to the low gravity environment of the International Space Station.  This drastic change in environmental conditions produced climbing behaviors not observed by these individuals on earth, reminiscent of distant relatives (e.g., \textit{Cephalotes goniodontus}), that proved effective in this new environment  \cite{countryman2015collective}.  Would a purely functional, optimized pavement ant design be able to exhibit such climbing behavior?  How can such capacity to adapt be described? 

Animal movement encodes information that is meaningfully interpreted by natural counterparts \cite{laidre2013animal}.  This is a behavior that roboticists are trying to replicate in artificial systems.  Can one artificial system successfully imitate another (possibly natural) system? Can a robotic system generate any arbitrary movement?
The success of this goal is often determined through comparison of behavioral landmarks inside tight, well-crafted contexts \cite{dragan2013legibility} or through comparison to rough data capture of human counterparts \cite{ames2014human} and other animals \cite{hopkins2009survey,haldane2013animal,cully2015robots}, with particular attention to the precision of end points of appendages \cite{ur52017,kinova2017,rethink2017,meijer2017performance} or rate of activity \cite{natureSpot}.  
Imitating the movement of biological organisms, using similar metrics for success, has been a topic in animation \cite{reynolds1999steering} and robotics \cite{egerstedt2005ants,powell2012motion}.  %, which is often dependent on the parameterization of the creature's movement.   
The review in \cite{breazeal2002robots} discusses ``robots that imitate humans'', describing ``very high fidelity playback.''  %How is this imitation defined?
In \cite{yamane2004synthesizing,nakaoka2004leg,gillies2009learning,kulic2012incremental,laviers2014ICCPS,joo2015panoptic}  motion capture data, a sparse representation, using a 10s or 100s of degrees of freedom, of human movement, seeds artificial system behavior.  On the other hand, models in animation are orders of magnitude more complex, using 1,000s and 10,000s of parameters \cite{zordan2004breathe,sueda2008musculotendon,sigal2010humaneva}, occluding clear bounds on imitation of natural behavior. 
Prior work in robotics \cite{knight2014expressive} and animation \cite{etemad2016expert} have termed motion ``expressive'' or ``affective'', which sets up a distinction between ``function'' and ``expression'' and is in contrast to work that shows the importance of context in resolving meaning in movement \cite{russell1987relativity,madi2017ICSR}.

The duality between function and expression in motion (analogous to function and form in product design) is discussed in the Laban/Bartenieff Movement System \cite{bartenieff,studd2013}.  Consider the same human motion in two distinct environments:  a living room and a jungle.  The agent \textit{thrashes its arms wildly, slashing at the air, and stomping its feet with heavy, sure-footed steps}.  In the living room, this behavior may express anger to a human viewer, but in the jungle, this motion is needed to accomplish the function of progressing through the heavy undergrowth of the jungle, which would be apparent to a human viewer.  This example establishes the idea that motion does not carry an inherent label, expression, or meaning even in the context of motion viewed by human counterparts.  Instead, this paper will pose that a greater variety of movement profiles increases the functionality and expressiveness of a platform.  %That is, the behavior of thrashing and simultaneously stomping described here is a behavior that may -- or may not -- be available to a moving platform, and the complexity of behaviors available is proportional to that platform's \textit{expressivity}.

This brings to bear a notion of expressivity in movement that is consistent with usage in computer science \cite{felleisen1991expressive}, genetics \cite{veltman2010understanding}, psychology \cite{restle1979coding}, and dance \cite{laviers2018choreographic}.
Information theory gives a clear model explaining how an 8-bit display, made of three LEDs, is more \textit{expressive} than a 1-bit display, made of one LED.  %Likewise, to create more characters in a font style, such as emoji, a more complex underlying ASCII representation is needed \cite{whitlock2016emoji}.  
Previous work in robotics \cite{blum1978power,donald1995information} has explored how much information is needed in tasks of \textit{sensing} the environment. 
Yet, quantitative bounds on capacity for \textit{actuation} of a moving platform (natural or artificial) to express information in the environment are not known; it is an open question as to whether it is possible for artificial systems to recreate the motion of natural systems \cite{elgin2010exemplification}.  

%This paper presents a measure for the capacity for motion complexity -- the expressivity -- of articulated platforms (both natural and artificial) and shows that this measure is stagnant and unexpectedly limited in extant robotic systems spanning the last 15 years. 
The paper points out the fundamental limitations on mechanization leveraging known limits on computation, previously proved by Turing \cite{turing1936computable}, challenging the idea that natural and artificial motion are comparable, and uses a proposed static information-theoretic expressivity measure to create observations analogous to Moore's Law \cite{schaller1997moore}, contextualizing the practical mechanical capacity of robots.  This analysis, applied to a variety of natural and artificial systems, shows trends \cite{mcmahon1983size,changizi2003relationship} in increasing capacity in both internal and external complexity for natural systems while artificial, robotic systems have increased significantly in the capacity of computational (internal) states but remained more or less constant in mechanical (external) state capacity.  This work shows that extant robots are not capable of the same multi-faceted behavior in rich, dynamic environments as complex natural systems and begins to quantify questions about the role of contextualized, redundant expression (as opposed to isolated, efficient function) in movement.

\section{Mechanization: An Ideal Process With Limits}

In \cite{turing1936computable} Turing outlines an \textit{a-machine}, a machine with a finitely complex mechanical head along an infinite tape where symbols can be stored.  The abstract machine requires the current configuration of the head, a list of basic instructions that tell the machine what to do in that configuration, and the complete configuration (state) of the whole thing.  
The components of an a-machine are given by the following list, loosely following \cite{sep-computability}:
\begin{itemize}
\item a finite set of $n$ machine states $Q=\{q_1,...,q_n\}$ ;
\item a finite set of $m$ symbols $\Sigma=\{\sigma_1,...,\sigma_m\}$ , e.g. $\Sigma=\{0,1,\epsilon\}$, where the result of machine computation, a computable number, is recorded in binary with a blank option, $\epsilon$
\item an infinite ``tape'' where these symbols are recorded, comprised of cells $c_1,c_2,c_3,...$, which is often pre-populated with a finite sequence of symbols that will set up for desired behavior when the machine is in operation;
\item current position along the tape, cell $c_h$, where $h\geq 1$;
\item a transition function $\delta : Q \times \Sigma \mapsto Q \times \Sigma \times \{-1,0,1\}$, which determines at a given state $q_i$ for a given scanned symbol $\sigma_i$ in $c_h$ how to update the position of the head $h$, i.e., it moves left, stays in place, or moves right.
\end{itemize}
Future work would introduce various instantiations of this idea, including essential pieces of the modern computer such as stored program architecture and clocking.  
However, these add-ons do not change a central premise of Turing's paper: the capacity of computing machines.  Specifically, he defines the class of numbers that can be computed by a properly formed (circle-free) machine to be enumerable (infinite but countable).  That is, there are many, many more numbers that cannot be computed (e.g., irrational numbers without formulas for computation like $\pi$) than can be.  This is seen through application of Cantor's diagonal process, which shows that the correspondence between natural numbers and computable numbers is one-to-one (or that the set of real numbers and computable numbers is not one-to-one) due to an inescapable recursive loop that traps an a-machine checking its own description number (this is known as the Halting Problem) \cite{petzold2008annotated}.  That is, we can imagine many more numbers than machines can compute, which means we can imagine many more machine behaviors than machines can perform.

Now, to establish a way of thinking about machine movement (mechanization), invert the a-machine, establishing an \ae-machine.  In this abstraction the idea of a physical workspace replaces Turing's idea of ``scratch paper'' where computations could be worked out.  An ideal mechanization machine will be able to complete tasks in the physical environment using extra workspace as needed.  This is similar to (but certainly not the same as\footnote{Indeed, unlike Turing, I want to motivate how this framework provides a \textit{limiting} picture of machines rather than a mechanistic model for how humans (may) work, as analysis in \cite{changizi2003relationship} also motivates.}) how a human artisan will use a workshop table during their work, placing part of a product off to the side while working on another element, using this tool or that to complete various steps, and increasing the size of their workshop as needed.  Similarly, this machine can perform actions inside its workspace, layering simple actions in sequence to produce desired effect on the environment.  Define such a machine as follows:
\begin{itemize}
\item a finite set of $n'$ states $Q'=\{q_1',...,q_n'\}$; 
\item a finite set of $m'$ actions, or \textit{motion primitives} $\Sigma'=\{\sigma_1',...,\sigma_m'\}$ , e.g. $\Sigma'=\{flexion,extension,\epsilon'\}$, where the result of machine mechanization is executed as either moving, moving in the opposite direction, or doing nothing, $\epsilon'$;
\item an infinite ``workspace'' where these actions are executed, comprised of cells $c_1',c_2',c_3',...$, which may be pre-populated with a finite set of primitives (or tools) that will set up for desired behavior when the machine is in operation;
\item current position in the workspace, cell $c_{h'}'$ ,where $h'\geq 1$;
\item a transition function $\delta' : Q' \times \Sigma' \mapsto Q' \times \Sigma' \times \{-1,0,1\}$, which determines at a given state $q_i'$ for a given motion primitive $\sigma_i'$ in $c_{h'}'$ how to update the position in the workspace $h$, which might be envisioned as a 1-, 2-, or 3-dimensional ``tape''.
\end{itemize}
  
What was a computation process (a sequence of logical symbols manipulated in an abstract, memory-like space) is now a mechanization process: a sequence of motion primitives executed in a discretized environment.  This sequence is likewise represented as a number -- one from Turing's set of computable numbers -- showing the infinite, but enumerable, action sequences possible to be executed by \ae-machines.  Thus, \ae-machines (idealized robots) have the same fundamentally limited capacity as a-machines (idealized computers).  That is, \textit{they cannot produce all the behaviors we might arbitrarily design}.  In particular, the cardinality of behaviors is equal to the cardinality of the set of computable numbers and is many orders of magnitude smaller than the cardinality of the set of real numbers. %(which is the mathematical ideal of all the behaviors we might arbitrarily imagine).
Moreover, just as Turing established subroutines to build his Universal Machine, we can create more complex behaviors of motion primitives that fire in sequence together, acting as a \textit{tool} in the workspace.  In practice, that tool could be ``software'' (a stereotyped, preprogrammed gesture or action) or ``hardware'' (an end effector attachment as a CNC machine selects distinct cutting tools).

\section{Static, Kinematic Capacity for A Mechanical Source}
\label{def}

This discrete way of thinking immediately sets up a practical formalism for the concept of \textit{expressivity}.  In order to consider {how complex the movement behaviors a machine can instantaneously exhibit are} (independent of environment or further augmentation), note that a bounding \textit{practical} limit on behavior is the cardinality of $Q'$ or $n'$.  While a robot with only a couple of degrees of freedom could split a given mechanization task into many steps, a robot with more might be able to do the entire task in one step\footnote{Imagine a switchboard of buttons, which may be arrange in a manner advantageous to a particular platform morphology.  In order to press one, two, three, four, and so on, buttons at the same time, a platform must have enough degrees of freedom (or the appropriate tool) to accomplish the task.}.  (We see a similar parallel in computational devices.)  Then, $n'$ is a good bottleneck measure on \textit{how complex} the instantaneous behavior of the machine can be (ignoring kinematically infeasible configurations and dynamic changes like velocity).
Thus, let $N$ be the number of actuator \textit{types} on a machine.  
Let $M_{i}$ be the number of degrees of freedom with a particular number of available configurations $R_i$, which is computed via counting from an actuators minimum to maximum range via its resolution where $i=1,...,N$.  
From this description of a machine's construction, let

\vspace{-.1in}
\begin{equation}
\mathcal{C}=\prod_{i=1}^N R_{i}^{M_i}
\end{equation}

\noindent be the number of discrete geometric or kinematic configurations (shapes) available to a platform.  
From there the \textit{kinematic expressivity} of that platform is the amount of information, or bits, needed to uniquely identify a configuration for that platform.  This is given by

\vspace{-.1in}
\begin{equation}
\mathcal{K}=\log_2(\mathcal{C}).
\end{equation}

On a robot with a simple gripper (which is either open or closed), two identical servos, {and a single LED $N=3$.  }  Assume each servo has $360^o$ range and $0.1^o$ resolution with a gripper that may be `open' or `closed' and an LED that may be `on' or `off', $R_1=3600$ with $M_1=2$ and $R_2=2$ with $M_2=2$. This becomes 

\vspace{-.1in}
$$2^2\times3600^2=5.2\times10^7 \mbox{ configurations.}$$  

\noindent This means the robot, as an information source, can express 

\vspace{-.1in}
$$\log_2(5.2\times10^7 \mbox{ configurations.})\approx 26\mbox{ bits}$$ 

\noindent of information, through state change, in its environment. Moreover, for this robot, most of it's complexity comes from \textit{mechanical} actuators.  Removing the LED from the analysis gives a \textit{kinematic mechanization capacity} of 

\vspace{-.1in}
$$\log_2(2^1\times3600^2=2.6\times10^7 \mbox{ configurations})\approx 25\mbox{ bits}.$$   

\noindent Thus, this simplistic robot can be compared to a 25-bit display in terms of \textit{mechanical} complexity.  

Any computer that is Turing-complete can, in theory, compute the set of computable numbers, given enough memory and time.  Similarly, a complete robot, with unlimited workspace and time, can produce a mechanization behavior associated with a computable number (i.e., it can simulate any \ae-machine).  However, the number of transistors in the CPU is a useful, implementation-specific measure (which has been growing \cite{schaller1997moore}) of precision to understand how practically powerful a given machine is.  Similarly, robots need more mechanical options in order to complete more complex mechanizations inside practical time and space limits.\footnote{This discrete view of robotic motion need not supplant the common practice of using a continuous vector-space model of configuration space \cite{choset2005principles}.  Indeed, both abstractions can be useful.}

\section{Computational and Mechanical Trends in Robotics}
\label{trends}
\begin{figure}[h!]
\centering
\includegraphics[width=\textwidth]{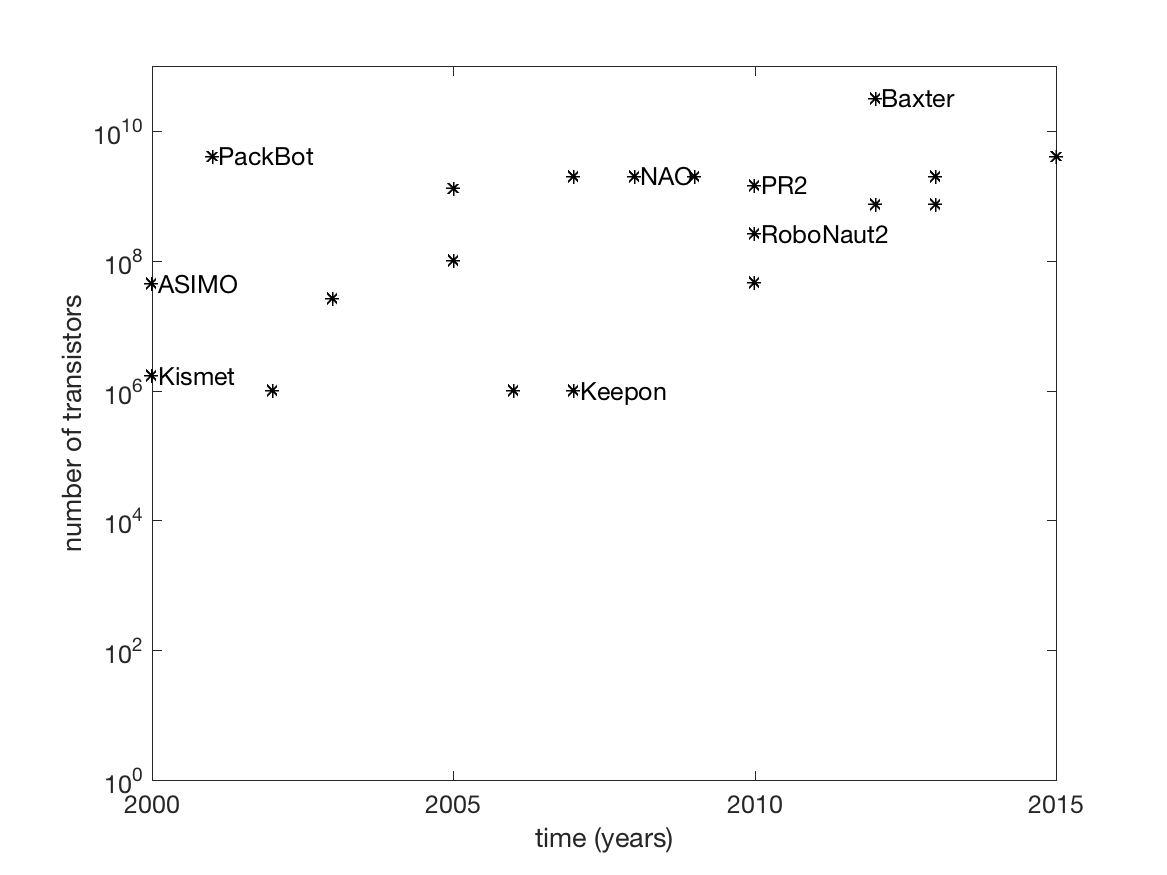}
\caption{The number of transistors, $t$, used in computational processors of robots over the last fifteen years.  The number of internal configurations available is then $2^{t}$. Some platform names are omitted for clarity; see Appendix for full list. }
\label{fplot1}
\end{figure}

This measure and way of thinking about the duality of computation and mechanization can categorize artificial systems. The plot in Figure \ref{fplot1} shows the number of transistors in the onboard CPU of a range of robots over the past 15 years.  Although, as many platforms are now internet connected, it is a limiting picture of the computational power available to these platforms.  Plots like this have been used to track the progress of computational power over time, which has roughly doubled every year, even serving as a driving goal for the industry (Moore's Law \cite{schaller1997moore}).  In such plots each additional component on an integrated circuit represents the ability to represent a larger -- or more precise -- number on a single chip.
Each new transistor adds a new power of 2 in representation precision.  Note, that the \textit{the number of transistors} in modern, stand-alone processors is in the billions.  To convert that to \textit{{possible machine configurations}}, where the actuators are transistors, the number 2 (which is the number of configurations for each actuator) has to be raised to that large number, resulting in a number of configurations that is bounded by $2^{10^{11}}$ or $10^{30102999566}$.  This number is an important, practical measure of how expressive the computers onboard robots are.

\begin{figure}[h!]
\centering
\includegraphics[width=\textwidth]{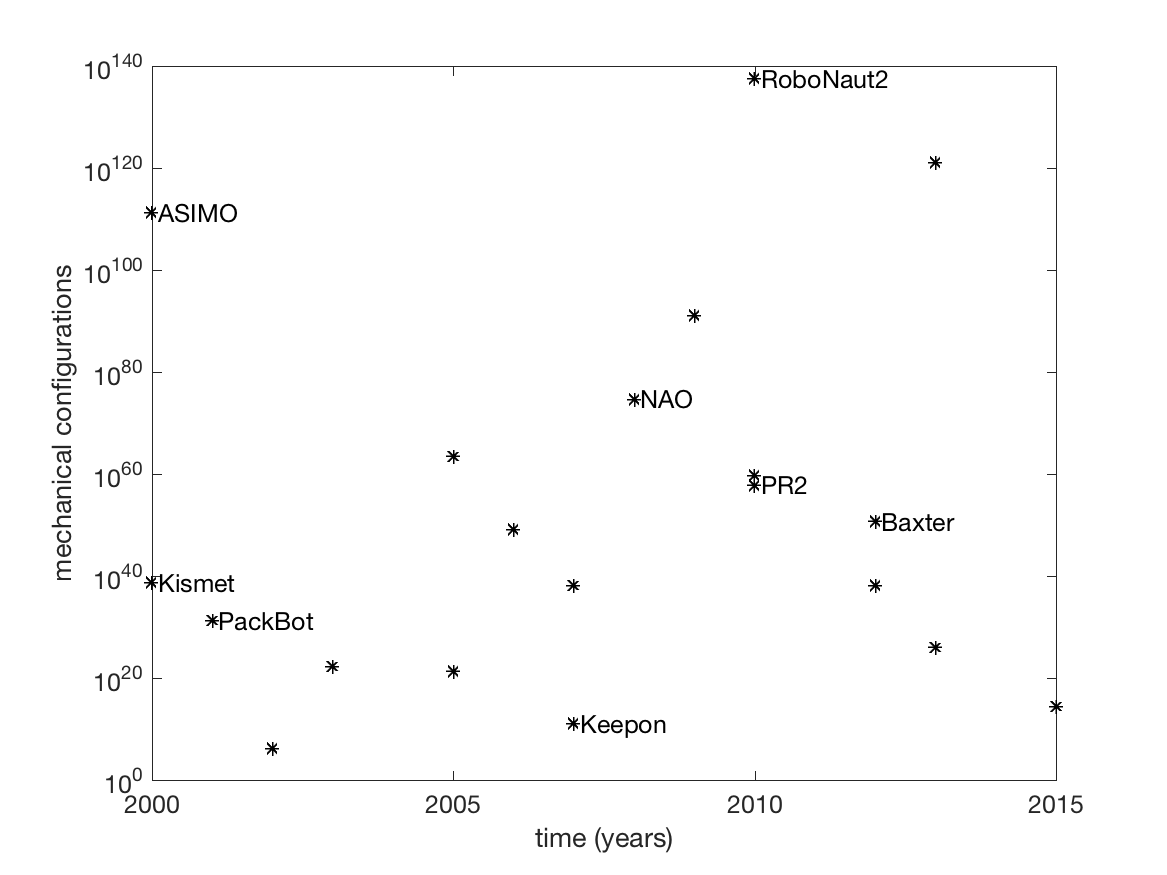}
\caption{The number of external configurations available, $\mathcal{C}$, for mechanization in robot platforms over time.  Several platforms have been assumed to have a positioning resolution of $0.1^{o}$ and an estimated range of motion. Some platform names are omitted for clarity; see Appendix for full list. }
\label{fplot2}
\end{figure}

Figure \ref{fplot2} shows how this expressivity measure has evolved on robots over time.  Specifically, the plot shows the number of possible kinematic configurations $\mathcal{C}$ for a number of rigid-body robots whose motion is governed by motors and encoders over time.  Like Moore's proxy of the number of transistors within a single chip, this kinematic configuration space is a static snapshot and does not account for dynamics, but it gives a starting point for measurement and comparison.  Here, the corresponding practical measure of how externally expressive these robots are is bounded by $10^{140}$.

\begin{figure}[h!]
\includegraphics[width=\textwidth]{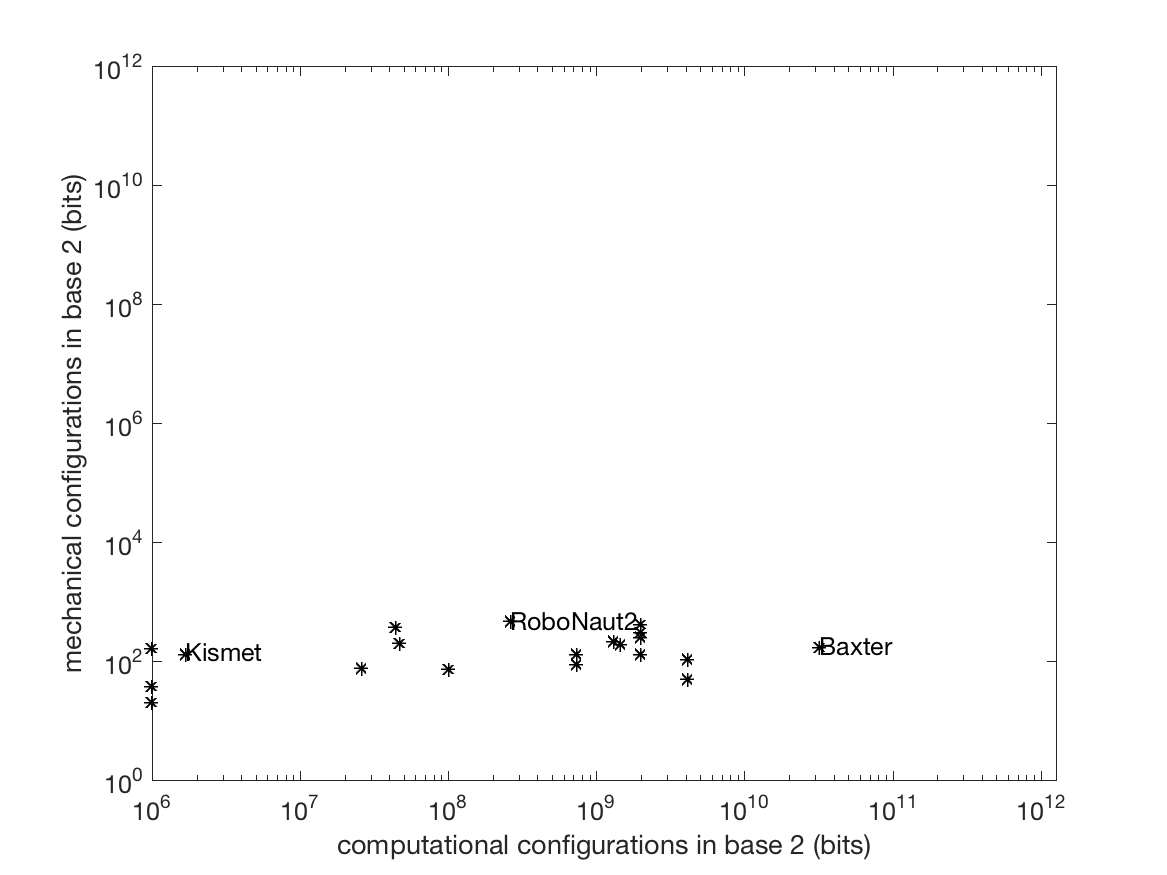}
\caption{A comparison of computational complexity (\# of transistors in the CPU) relative to mechanical complexity ($\mathcal{K}$) on robotic platforms over time. Platform names are omitted for clarity; see Appendix for full list.  The unit of measure on both axes is \textit{bits}; a square plot highlights imbalance between internal and external states.}
\label{fplot3}
\end{figure}

By converting the number of configurations $\mathcal{C}$ to a number represented in a base 2 number system, $\mathcal{K}$, the change in the computational capacities of these platforms can be compared to their mechanization capacities as in Fig. \ref{fplot3}.  
%To manage the large numbers, each capacity is recorded as the logarithm, in base 2, of the total number of available configurations.  
This log-log plot provides a comparison in terms of the number of bits which it would take to describe the largest number that would fit in the onboard CPU versus the number of bits needed to uniquely represent each pose.  The plot shows a dramatic imbalance between computation and mechanization capacities and a flat trend in the order of magnitude of mechanical capacity over the past 15 years.   
Through this lens, the NAO Aldebaran, a small humanoid robot is comparable to a 1960s computer chip with only 256 transistors (see Appendix for detailed calculation).  

%%%% SUBFIGURE SOURCE

%\begin{figure}
%    \centering
%    \begin{subfigure}[b]{0.49\textwidth}
%        \includegraphics[width=\textwidth]{plot2}
%        \caption{}
%        \label{fplot1}
%    \end{subfigure}
%    ~ %add desired spacing between images, e. g. ~, \quad, \qquad, \hfill etc. 
%      %(or a blank line to force the subfigure onto a new line)
%    \begin{subfigure}[b]{0.49\textwidth}
%        \includegraphics[width=\textwidth]{plot1}
%                \caption{}
%        \label{fplot2}
%    \end{subfigure}
%    %add desired spacing between images, e. g. ~, \quad, \qquad, \hfill etc. 
%    %(or a blank line to force the subfigure onto a new line)
%%    \begin{subfigure}[b]{0.33\textwidth}
%%        \includegraphics[width=\textwidth]{plot3}
%%        \label{fplot3}
%%    \end{subfigure}
%    \caption{A comparison of internal and external robot states over time. Some platform names are omitted for clarity; see Appendix for full list. (a) The number of transistors, $t$, used in computational processors of robots over the last fifteen years.  The number of internal configurations available is then $2^{t}$.  (b) The number of external configurations available for mechanization in robot platforms over time.  Several platforms have been assumed to have a positioning resolution of $0.1^{o}$ and an estimated range of motion.}\label{2plots1}
%\end{figure}

\section{A Comparison Between Machines and Animals} \label{biocomp}
The previous section outlines a way to compare robot capacity for complex behavior, but the same method can be applied to biological creatures -- with the caveat that such systems may involve processes not extant in today's machines.  Processes like ``mechanization'' and ``computation'' are not appropriate labels for natural systems, and the limit on behavior so carefully derived by Turing is not known to apply here. That is, the cardinality of the set of possible behavior of natural systems may (or may not) be larger than the cardinality of the set of computable numbers.  However, we can loosely compare ``computation'' to internal state changes -- those that do not meaningfully impact the environment -- and ``mechanization'' to external state changes -- those that cause change in the environment.  Moreover, the difference between sensing and actuation blurs for these systems where movement is tightly linked to sensing, e.g. fovea and hairs.

In prior work, the movement of a \textit{C. Elegan} was analyzed using a curve parameterized by 100 angles; then, after capturing an extended period of behavior on laboratory agar, a principle component analysis was performed, revealing that the structure of the behavior could be explained as a superposition of four primary poses \cite{stephens07}.  Observing the animal through this lens provided new insights into the behavior of this well-studied animal \cite{stephens08a,stephens08b}, implying a meaningful parameterization, further explored in \cite{gomez2016hierarchical}.  
Stephen's model can be used to compare the \textit{C. Elegan} to modern robots.
A \textit{C. Elegan} has 302 neurons, which can be approximated to be either `firing' or `not' in a static snapshot of time.  Then, each of the 100 angles have a typical range, shown in the empirical results in \cite{stephens07}, between $ -1.5$ radians to $1.5$ radians, implying $0.1$ radians of precision.  Thus, using the metric proposed in Section \ref{def}, the kinematic mechanization capacity for this model of a simple animal can be calculated as 

\vspace{-.2in}
\begin{eqnarray}
\mathcal{K}=\log_2(30^{100})
=\log_2(5.2\times10^{147}\mbox{ configurations})\approx491\mbox{ bits}.
\end{eqnarray}

\noindent The linear model derived in \cite{stephens07} can also be used to derive a slightly more compact expression.  Here, the idea is that it's possible that the organism, in a particular task or environment, does not utilize the full span of motion modeled by 100 angles.  This could be a function of behavioral patterning, environment, or experimental set up\footnote{However, like the pavement ant, the larger behavioral space intoned by the anatomical model may be used in future, unforeseen environments.}.  Here, linear combinations four ``eigenworms'' describe 95\% of the observed worm behavior.  The weights on each eigenworm, $\alpha_1$, $\alpha_2$, $\alpha_3$, and $\alpha_4$, also have an observed range, which can be estimated from \cite{stephens07} as $\alpha_1: [-2,2]$, $\alpha_2: [-2,2]$, and $\alpha_3: [-5,5]$.  Four the fourth, which was not found to have behavioral significance and is not provided in \cite{stephens07}, assume $\alpha_4: [-2,2]$.  For each, assume $0.1$ in precision.  This leads to the following calculation:

\vspace{-.2in}
\begin{eqnarray}
\mathcal{K}=\log_2(40^{3}\times100^{1})
=\log_2(6.4\times10^{6}\mbox{ configurations})
\approx23 \mbox{ bits}.
\end{eqnarray}

\begin{figure}[h!]
\centering
\includegraphics[width=\columnwidth]{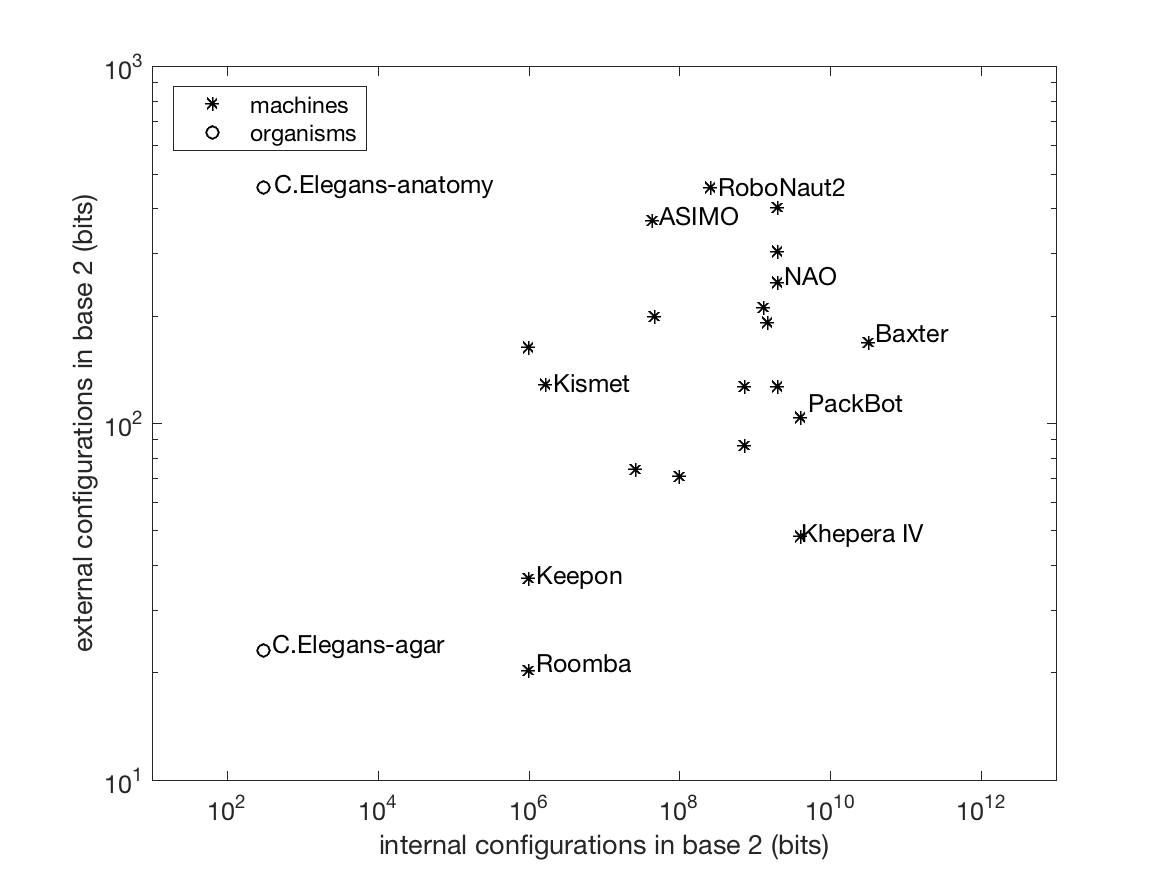}
\caption{The machine data points are the same as those in Fig. \ref{fplot2}.  The \textit{C. Elegan} x-axis points are based on the number of neurons (302). The y-axis data points plotted for \textit{C. Elegans} show well-established behavioral models, one based on anatomy and the other based on exhibited behavior on agar \cite{stephens07}.}
\label{fplot2ce}
\end{figure}

The summary of this analysis is plotted on Fig. \ref{fplot2ce}.  The plot gives a sense of the capacity of today's robots.  It is shown that \textit{C. Elegans} are apt natural analogs for the mechanical capacity of many robots.  This does not mean that \textit{C. Elegans} can do any of the specific functions that these robots can do; it means that they can do the same number things at the same time -- or are as {expressive}. This is consistent with the fact that today's robots are created for single-tasks in limited environments.  \textit{C. Elegans} must forage for food and carry out other critical to life tasks in real, dynamic environments\footnote{There are distinct Reynolds numbers at play.  Inertial forces dominate the selected robots, while viscous forces dominate the motion of \textit{C. Elegans}; however, this analysis is not considering forces, only the complexity of behavioral snapshots.}.  

\begin{figure}[h!]
\centering
\includegraphics[width=\columnwidth]{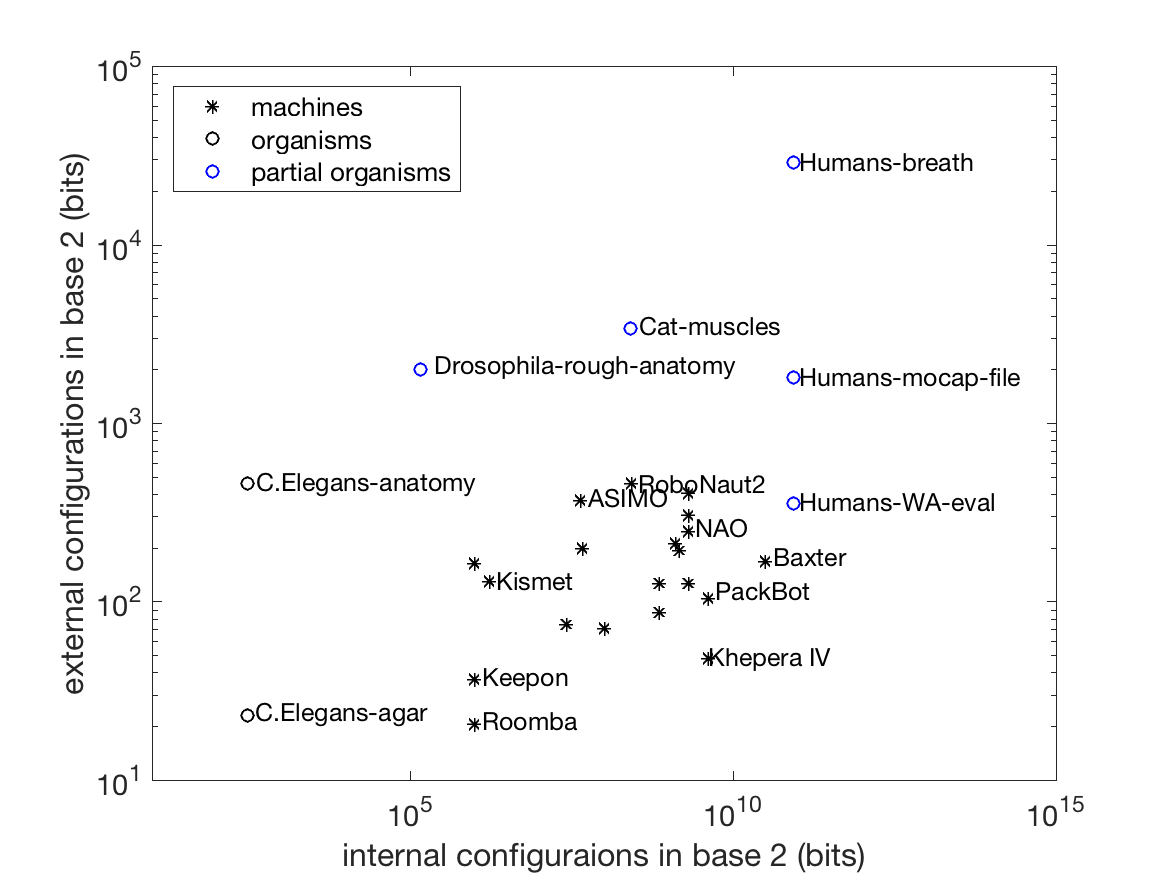}
\caption{The data in Fig. \ref{fplot2ce} appended with data of a few more natural systems.  The x-axis values for both of these organisms are based on the number of neurons \cite{azevedo2009equal}; the y-axis values are given by: Drosophila-rough-anatomy: analysis of  \cite{chyb2013atlas}; Cat-muscles: generalizing the model of muscles in \cite{ting2005limited}; Humans-WA-eval: the structure of \cite{wsu-eval}; Humans-mocap: the structure of a Natural Point OptiTrack motion capture file; and Humans-breath: the simulation in \cite{zordan2004breathe}. See Appendix for more detail.}
\label{fplot2ce2}
\end{figure}

Such detailed motion models for other animals are not common, but a few clear vignettes are appended to Fig \ref{fplot2ce} in Fig. \ref{fplot2ce2}.  Relevant work done on fruit flies (\textit{Drosophila}) abstracts their state as a vector with a heading over time \cite{dankert2009automated}, but an anatomical analysis via \cite{chyb2013atlas} can estimate instantaneous complexity of the organism.  Researchers look to muscle activity involved in postural control for cats with weights having precision to the hundredth decimal place \cite{ting2005limited}; generalizing this precision to all estimated 517 \cite{sebastiani2005mammalian} muscles in cats gives an estimate of the complexity of their muscular control, leaving out the mechanical advantages of their compliant bones, skin, and hair.  For human motion, three different established guides to behavior (a public health diagnostic \cite{wsu-eval}, a Natural Point OptiTrak motion capture file, and an animation of breathing \cite{zordan2004breathe}) of body posture can be used to show similarity between these models and those that identified behavioral understanding of \textit{C. Elegans}.  Either, humans are as externally complex as \textit{C. Elegans} or more detailed models are needed.  

\section{Conclusions}
The paper has shown a relationship between mechanization and computation, pointing to a fundamental limit on machine behavior.  The paper has introduced a static, practical measure for expressivity, clarifying prior points of view on function versus expression.  Finally, the paper uses this work to compare extant artificial systems to natural systems, showing many modern robots, including humanoids, are about as expressive as a microscopic worm. This work is an essential piece of information in the discussion, which is by now mainstream, around the effect of machine-based automation of human tasks, particularly in manufacturing.  Humans operate in dynamic environments where variability in movement is essential in accommodating and coping with an unpredictable world.  Robots do well in repeatable tasks where environmental factors are controlled (as in a factory), but do not have the same capacity to adapt.  The trends pointed to here in robotics may help guide system development:  in hardware, soft and compliant robots, and, in software, increasing motion variability will offer robotic systems greater expressivity. Further, considering the wide, possibly unused, capacity for motion of natural systems, may produce improved understanding of how animals adapt both as individuals and as species.

\section*{Acknowledgments}
This work was funded by DARPA award  \#D16AP00001.  Undergraduate students Jialu Li and Varun Jain helped collect and estimate data on each artificial platform reviewed.  %Dr. Joe Rokicki read an early version of the draft, providing insightful comments to help with the clarity of the explanations.  Drs. Todd Murphy and William Bialek gave particularly helpful feedback via high-level conversations.

%%%%%%%%%%%%%%%%%%%%%%%%%%%%%%%%%%%%%%%%%%
%% optional
\appendixtitles{yes} %Leave argument "no" if all appendix headings stay EMPTY (then no dot is printed after "Appendix A"). If the appendix sections contain a heading then change the argument to "yes".
%\appendixsections{multiple} %Leave argument "multiple" if there are multiple sections. Then a counter is printed ("Appendix A"). If there is only one appendix section then change the argument to "one" and no counter is printed ("Appendix").

%\newpage
\appendix
\section{Platform Details} \label{appen}
Additional numbers used to calculate data points in Figures \ref{fplot1}-\ref{fplot3}.  For mechanical configurations, see below.  Many of these were determined through observation and are meant to indicate the visual expressiveness of the platforms.  Many rows represent multiple, homogeneous degrees of freedom (indicated through `l/r' for `left' and `right' and with numbers in parenthesis) for space.

\begin{table}[h!] 
\begin{center}
\begin{tabular}{|p{.32\columnwidth}||p{.3\columnwidth}||p{.08\columnwidth}|}
\hline
\textbf{Robot \& DOF} & \textbf{Range / Resolution} & \textbf{$R_i$}\\
  \hline \hline
    NAO &  &  \\
  \hline
      l/r hand & open/close & 2 \\
  \hline
    head yaw & -119.5 to 119.5 / .1 & 2390 \\
  \hline
    head pitch (at 0 yaw) & -38.5 to 29.5 / .1 & 680 \\
  \hline
    l/r shoulder pitch & -119.5 to 119.5 / .1 & 2390 \\
  \hline
      l/r shoulder yaw & -119.5 to 119.5 / .1 & 2390 \\
  \hline
      l/r shoulder roll & -88.5 to -2 / .1 & 865 \\
  \hline
      l/r wrist yaw & -104.5 to 104.5 / .1  & 2090 \\
  \hline
      pelvis & -65.6 to 42 / .1 & 1076 \\
  \hline
      l/r hip roll & -21.7 to 45.2 / .1 & 669 \\
  \hline
      l/r hip pitch & -88 to 27.7 / .1 & 1157 \\
  \hline
      l/r knee pitch & -5.3 to 121.0 / .1 & 1263 \\
  \hline
      l/r ankle pitch & -68.2 to 52.9 / .1 & 1211 \\
  \hline
      l/r ankle roll & -22.8 to 44.1 / .1 & 669 \\
  \hline
  Baxter &  &  \\
  \hline
  l/r S1 & open/close & 2 \\
  \hline
  l/r E1 & -2.864 to 150 / 0.1& 1530 \\
  \hline
    l/r W1 &-90 to 120 / 0.1& 2100 \\
  \hline
l/r S0 & -97.5 to 97.5 / 0.1 & 1950 \\
  \hline
l/r E0 &-175.0 to 175.0 / 0.1& 3500 \\
  \hline 
l/r W0 & -175.25 to 175.25 / 0.1 & 3505 \\
  \hline
l/r W2&-175.25 to 175.25 / 0.1 & 3505 \\
\hline
    Khepera IV & & \\
  \hline
l/r wheel & 360 / 0.1 / .1 & 3600 \\
  \hline
    Roomba & & \\
  \hline
l/r wheel & 360 / 0.1 / .1 & 3600 \\
  \hline
Kismet &  &  \\
  \hline
l/r ears pitch & -67.5 to 67.5 / 0.1  & 1350 \\
  \hline
l/r ears yaw & -22.5 to 22.5 / 0.1& 450 \\
  \hline
l/r eyelids &-1.5 to 1.5 / 0.1 & 30 \\
  \hline
l/r brows pitch & -10 to 10 / 0.1 & 200 \\
  \hline
l/r lips & -30 to 30 / 0.1 & 600 \\
  \hline
jaw & -22.5 to 22.5 / 0.1& 450 \\
  \hline
PR2 & & \\
  \hline
l/r shoulder pan& 170 / 0.1& 1700 \\
  \hline
l/r shoulder tilt&115 / 0.1 & 1150 \\
  \hline
l/r upper arm roll& 270 / 0.1 & 2700 \\
  \hline
l/r elbow flex &140 / 0.1 & 1400 \\
  \hline
l/r forearm roll & 360 / 0.1& 3600 \\
  \hline
l/r wrist pitch &130 / 0.1& 1300 \\
  \hline
l/r wrist roll&360 / 0.1  & 3600 \\
  \hline
head pan& 350 / 0.1& 3500 \\
  \hline
head tilt&115 / 0.1& 1150 \\
  \hline
Big Dog &  & \\
  \hline
each leg (5) (x4) & 150 / 0.08 & 1875 \\
  \hline
        \end{tabular}
\end{center}
\end{table}

\begin{table}[h!] 
\begin{center}
\begin{tabular}{|p{.32\columnwidth}||p{.3\columnwidth}||p{.08\columnwidth}|}
\hline
\textbf{Robot \& DOF} & \textbf{Range / Resolution} & \textbf{$R_i$}\\
  \hline \hline
ASIMO & & \\
  \hline
head (3)& 150 / 0.08& 1875 \\
  \hline    
arms (14) & 150 / 0.08& 1875 \\
  \hline
hands (4)&150 / 0.08 & 1875 \\
  \hline
torso (1) & 150 / 0.08& 1875 \\
  \hline
legs (12) & 150 / 0.08& 1875 \\
  \hline
Little Dog & &\\
  \hline
l/r front knee RY&-177 to 57 / 0.1& 2340 \\
  \hline
l/r front hip RX& -34 to 34 / 0.1  & 680 \\
  \hline
l/r front hip RY& -200 to 137 / 0.1& 337 \\
  \hline
l/r back knee RY&-57 to177 / 0.1 & 2340 \\
  \hline
l/r back hip RX & -34 to 34 / 0.1 & 680 \\
  \hline
l/r back hip RY & -137 to 200 / 0.1 & 337 \\
  \hline
Robotnaut2 & & \\
  \hline
head yaw/pitch/roll  &150 / 0.08 & 1875 \\
  \hline
l/r hands (12) & 150 / 0.08 & 1875 \\
  \hline
l/r arms (7) & 150 / 0.08 & 1875 \\
  \hline
KeepOn &  &  \\
  \hline
tilt &-40 to 40 / 0.08& 1000 \\
  \hline
pan& -180 to 180 / 0.08 & 4500 \\
  \hline
pon & 0 to 100 / 0.08 & 1250 \\
  \hline
side & -25 to 25 / 0.08 & 625 \\
  \hline
RoboSapien&  &  \\
  \hline
l/r elbows & -90 to 90 / 0.1 & 1800 \\
  \hline
l/r shoulders & -30 to 150 / 0.1 & 1800 \\
  \hline
torso & -67.5 to 67.5 / 0.1 & 1350 \\
  \hline
l/r hips & -60 to 60 / 0.1 & 1200 \\
        \hline
Darwin &  &  \\
  \hline
neck pitch& -25 to 25 / 0.1 & 500 \\
  \hline    
neck roll& -90 to 90 / 0.1 & 1800 \\
  \hline
l/r elbow & 0 to 150 / 0.1 & 1500 \\
  \hline
l/r shoulder rotation & -100 to 100 / 0.1 & 2000 \\
  \hline
l/r shoulder compression & -15 to 15 / 0.1 & 300 \\
  \hline
l/r knee & 0 to 150 / 0.1 & 1500 \\
  \hline
l/r foot& 0 to 90 / 0.1 & 900 \\
  \hline
l/r waist rotation & -15 to 15 / 0.1  & 300 \\
  \hline
l/r knee/foot & -75 to 75 / 0.1 & 1500 \\
  \hline
l/r waist bend & 0 to 100 / 0.1 & 1000 \\
  \hline
  Aibo &  &  \\
  \hline
head pan & -89 to 89 / 0.1 & 1780 \\
  \hline
head tilt& -62.5 to 62.5 / 0.1& 1250 \\
  \hline
head roll& -29 to 29 / 0.1 & 580 \\
        \hline
shoulders (4) & 0 to 100 / 0.1 & 1000 \\
  \hline
torso & -117 to 117 / 0.1& 2340 \\
  \hline    
knees (4)& 0 to 175 / 0.1 & 1750 \\
  \hline
l/r ears &0 to 20 / 0.1 & 200 \\
  \hline
tail (front to back) & -22.5 to 22.5 / 0.1 & 450 \\
  \hline
tail (left to right) & -12.5 to 12.5 / 0.1 & 250 \\
  \hline
      \end{tabular}
\end{center}
\end{table}

\begin{table}[h!] 
\begin{center}
\begin{tabular}{|p{.32\columnwidth}||p{.3\columnwidth}||p{.08\columnwidth}|}
\hline
\textbf{Robot \& DOF} & \textbf{Range / Resolution} & \textbf{$R_i$}\\
  \hline \hline
Packbot &  &  \\
  \hline
shoulder rot. & 0 to 360 / 0.1& 3600 \\
  \hline
shoulder pivot&0 to 220 / 0.1& 2200 \\
  \hline
E1 pivot & 0 to 340 / 0.1& 3400 \\
  \hline
E2 pivot& 0 to 340 / 0.1& 3400 \\
  \hline
gripper rot.&0 to 360 / 0.1& 3600 \\
  \hline
gripper I/O & 180 / 0.1  & 1800 \\
  \hline
head rot. & 0 to 360 / 0.1& 3600 \\
  \hline
flipper& 0 to 360 / 0.1& 3600 \\
  \hline
Simon & &\\
  \hline
torso (2) &-75 to 75 / 0.1 & 1500 \\
  \hline
l/r arm (7) & 0 to 200 / 0.1 & 2000 \\
  \hline
face (5)&0 to 200 / 0.1 & 2000 \\
  \hline
Cheetah & &\\
  \hline
hip rot. (4)& 0 to 30 / 0.1 & 300 \\
  \hline
  hip (4)& 0 to 150 / 0.1& 1500 \\
  \hline
knee (4) & 0 to 200 / 0.1 & 2000 \\
  \hline
spine& -10 to 10 / 0.1 & 200 \\
  \hline
LBR iiwa & &\\
  \hline
axis 1& -170 to 170 / 0.1& 3400 \\
  \hline
axis 2& -120 to 120 / 0.1& 2400 \\
  \hline
axis 3& -170 to 170 / 0.1& 3400 \\
  \hline
axis 4& -120 to 120 / 0.1& 2400 \\
  \hline
axis 5& -170 to 170 / 0.1& 3400 \\
  \hline
axis 6& -120 to 120 / 0.1& 2400 \\
  \hline
axis 7& -175 to 175 / 0.1& 3500 \\
  \hline
KR60HA& & \\
  \hline    
axis 1& -185 to 185 / 0.1& 3700 \\
  \hline
axis 2&-135 to 35 / 0.1& 1700 \\
  \hline
axis 3& -120 to 158 / 0.1& 1780 \\
  \hline
axis 4& -350 to 350 / 0.1& 7000 \\
  \hline
axis 5& -119 to 119 / 0.1& 2380 \\
  \hline
axis 6& -350 to 350 / 0.1& 7000 \\
  \hline
\end{tabular}
\end{center}
\end{table}

\clearpage
\newpage
For computational configurations, the following values were used. Here, $\mathcal{C}$ is $2^x$ where $x$ is the number of transistors.  Indeed, often, another, larger computer (or cluster of processors) is networked to these machines through wireless or wired connections.  But, it is instructive nonetheless to compare how much more sophisticated the computational power (even that which is on board) is relative to the mechanical power.

\begin{table}[h!] 
\begin{center}
\begin{tabular}{|p{.2\columnwidth}||p{.3\columnwidth}||p{.2\columnwidth}|}
\hline
\textbf{Robot} & \textbf{Processor} & \textbf{\# of transistors}\\
  \hline \hline
NAO & Atom Z530  & 4.7E+07\\
  \hline
Baxter &  3rd Gen Intel Core i7-3770  & 1.40E+09\\
  \hline
      Khepera IV &  ARM Cortex-A8  & 2.00E+09\\
  \hline
    Roomba & & 1.00E+06\\
  \hline
Kismet & Motorola 68332 (4) &  1.68E+06\\
  \hline
PR2 & Two Quad-Core i7 Xeon  (8 cores) & 1.462E+09 \\
  \hline
Big Dog & Pentium CPU & 1.30E+09\\
  \hline
ASIMO & Pentium III-M 1.2 GHz & 4.40E+07\\
  \hline
Little Dog & Pentium CPU & 2.00E+09\\
  \hline
Robotnaut2 & & 2.622E+08 \\
  \hline
KeepOn & PS234 & 1.00E+06 \\
  \hline
RoboSapien& 200MHz ARM9 & 2.60E+07 \\
  \hline
Darwin & Intel Atom Z510 & 4.70E+07 \\
  \hline
Aibo & 64 bit RISC  & 1.00E+06 \\
  \hline
Packbot & Pentium 3 & 4.50E+07 \\
  \hline
Simon & & 2.00E+09\\
  \hline
Cheetah & & 7.31E+08\\
  \hline
LBR iiwa & & 7.31E+08\\
  \hline
KR60HA& & 1.00E+08\\
  \hline    
\end{tabular}
\end{center}
\end{table}

For the natural systems analyzed, the following values were used based on \cite{stephens07,chyb2013atlas,ting2005limited,sebastiani2005mammalian,wsu-eval,zordan2004breathe}. 

\begin{table}[h!] 
\begin{center}
\begin{tabular}{|p{.32\columnwidth}||p{.3\columnwidth}||p{.08\columnwidth}|}
\hline
\textbf{Organism \& DOF ($M_i$)} & \textbf{Range / Resolution} & \textbf{$R_i$}\\
  \hline \hline
\textit{C. Elegan} (anatomy) &   & \\
  \hline
$\theta$ (100) &  -1.5 rad to 1.5 / 0.1  & 150\\
  \hline
\textit{C. Elegan} (agar behavior) &  &\\
  \hline
    $\alpha_1$,  $\alpha_2$, and $\alpha_4$   & -2 to 2 / 0.1 & 40 \\
  \hline
    $\alpha_3$   & -5 to 5 / 0.1 & 100 \\
  \hline
\textit{Drosophila}  &  &\\
  \hline
Tarsus 5 (6) & 0 to 180$^o$ / 0.1$^o$ & 1800\\
  \hline
Tarsus 4 (6) & 0 to 180$^o$ / 0.1$^o$ & 1800\\
  \hline
Tarsus 3 (6) & 0 to 180$^o$ / 0.1$^o$ & 1800\\
  \hline
Tarsus 2 (6) & 0 to 180$^o$ / 0.1$^o$ & 1800\\
  \hline
  Tarsus 1 (6) & 0 to 180$^o$ / 0.1$^o$ & 1800\\
  \hline
  Tibia (6) & 0 to 180$^o$ / 0.1$^o$ & 1800\\
  \hline
Femur (6) & 0 to 180$^o$ / 0.1$^o$ & 1800\\
  \hline
  Trochanter (6) & 0 to 360$^o$ / 0.1$^o$ & 3600\\
  \hline
  Coxa (6) & 0 to 10$^o$ / 0.1$^o$ & 100\\
  \hline
  Wing cells (12)) & flexed or relaxed & 2\\
  \hline
  Wing hinge (6) & 0 to 180$^o$ / 0.1$^o$ & 1800\\
  \hline
Halteres (6) & 0 to 360$^o$ / 0.1$^o$ & 3600\\
  \hline
  Head, Thorax, Abdomen (9) & 0 to 45$^o$ / 0.1$^o$ & 450\\
  \hline
  Probiscis & in or out & 2\\
  \hline
  Antennae (12) & 0 to 10$^o$ / 0.1$^o$ & 100\\
  \hline
  Bristles (200) & flexed or not & 2\\
  \hline
  Hairs (1000) & flexed or not & 2\\
  \hline
      \end{tabular}
\end{center}
\end{table}

\begin{table}[h!] 
\begin{center}
\begin{tabular}{|p{.32\columnwidth}||p{.3\columnwidth}||p{.08\columnwidth}|}
\hline
\textbf{Organism \& DOF ($M_i$)} & \textbf{Range / Resolution} & \textbf{$R_i$}\\
  \hline \hline
Cat &  &\\
  \hline
Muscles (517) & 0 to 1 / 0.01 & 100\\
  \hline
Human (WA-eval) &  &\\
  \hline
Each listed diagnostic  \cite{wsu-eval} (37) & various / 0.1 & various \\
  \hline
Human (mocap) &  &\\
  \hline
DOFs (66) & -180$^o$ to 180$^o$ / 0.000001 & 360000000\\
  \hline
Human (breath) &  &\\
  \hline
Muscle-spring elements (1500) & -1 to 1 / 0.000001 & 1000000\\
  \hline    
\end{tabular}
\end{center}
\end{table}

%% Use plainnat to work nicely with natbib. 

%=====================================
% References, variant B: external bibliography
%=====================================
\newpage
\externalbibliography{yes}
\section*{References}
\bibliography{papers}

\newpage

\appendix
\section{Applied Examples}
\label{examples}
In this section, the measure introduced in Section \ref{def} will be applied to some instructive examples: a small humanoid robot and Vegas's Bellagio fountains.  Passive systems like the motion of a falling leaf versus a falling brick might also be considered.  %\textcolor{red}{\textit{Other examples:  a leaf, soccer player disguising intent,....}}

\subsection{Aldebaran NAO Humanoid Robot} \label{NAO_ex}
%The Aldebaran NAO robot is a captivating machine.  Indeed, advances in feedback control and robotics were needed to build it.  Appropriately, there is something impressive about seeing it move.  The computation provided in this section, however, may call into question how useful it can be for replicating human behavior in any context.  

\begin{figure}[h!]
\centering
\includegraphics[width=.9\textwidth]{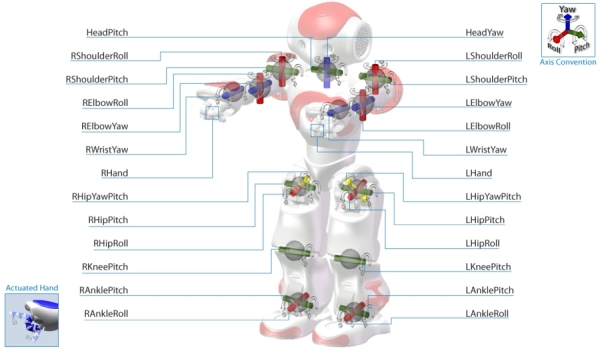}
\caption{A diagram which lists the degrees of freedom on an Aldebaran NAO robot.  In addition to these mechanical degrees of freedom the platform contains an ATOM Z530 onboard computer processor, which has 47 million transistors on board. \cite{G2009,aldebaran2009robotics}} 
\label{NAO}
\vspace{-.1in}
\end{figure}

Figure \ref{NAO} and Table \ref{nao_dof} outline the basic capabilities of the platform where the sensor resolution (an encoder with $0.1^o$ precision) has been used to determine $R_i$.

\begin{table}[h!] 
\begin{center}
\begin{tabular}{|p{.3\columnwidth}||p{.3\columnwidth}||p{.1\columnwidth}|}
\hline
{DOF} & {Range / Resolution} & \textbf{$R_i$}\\
  \hline \hline
    l/r hand & open/close & 2 \\
  \hline
    head yaw & -119.5 to 119.5 / .1 & 2390 \\
  \hline
    head pitch (at 0 yaw) & -38.5 to 29.5 / .1 & 680 \\
  \hline
    l/r shoulder pitch & -119.5 to 119.5 / .1 & 2390 \\
  \hline
      l/r shoulder yaw & -119.5 to 119.5 / .1 & 2390 \\
  \hline
      l/r shoulder roll & -88.5 to -2 / .1 & 865 \\
  \hline
      l/r wrist yaw & -104.5 to 104.5 / .1  & 2090 \\
  \hline
      pelvis & -65.6 to 42 / .1 & 1076 \\
  \hline
      l/r hip roll & -21.7 to 45.2 / .1 & 669 \\
  \hline
      l/r hip pitch & -88 to 27.7 / .1 & 1157 \\
  \hline
      l/r knee pitch & -5.3 to 121.0 / .1 & 1263 \\
  \hline
      l/r ankle pitch & -68.2 to 52.9 / .1 & 1211 \\
  \hline
      l/r ankle roll & -22.8 to 44.1 / .1 & 669 \\
  \hline
\end{tabular}
\end{center}
\caption{NAO Aldebaran robot, mechanical degrees of freedom.} \label{nao_dof}
\end{table}

Thus, the kinematic mechanization capacity is calculated as
\begin{eqnarray}
\mathcal{K}=\log_2(2^2\times2390^5\times680^1\times940^2\times865^2...\times2090^2
\times1076^1\times669^4\times1157^2\times1263^2\times1211^2)\\\nonumber
=\log_2(4.1\times10^{71}\mbox{ configurations})\approx238 \mbox{ bits}
\end{eqnarray}

This calculation includes physical combinations which are kinematically or dynamically infeasible.   Further, the capacity for changes in motor speed between configurations is not reflected here and would increase the complexity of behavior. Thus, the number may be seen as a static measure.

\subsection{Bellagio Water Fountains} 
\label{bellagio_ex}
Consider a tourist attraction, like the Bellagio water fountains in Las Vegas, NV.  Tourists line up every hour to watch this famous display, routinely included in lists of popular Vegas attractions.  This is to say that the fountain display is visually very interesting, or expressive, for human watchers.  %Let's compare how much more complex it is than typical robots and consider how much less complex it is than most computers via the proposed measure.
The fountain has about 1,200 water cannons with 5,000 lights as part of its display.  It also has the ability to create fog and features popular or famous music during the shows.  For this analysis \cite{vegas1,vegas2}, we'll consider only the water cannons and lights. The cannons come in four types: robotic Oarsman and three sizes of Shooters.  The 208 Oarsman are articulated cannons with active control; the Shooters simply blast water at three predetermined pressure settings, each having a single pressure setting according to their size.  The lights can be a range of colors.

\begin{table}[h!] 
\begin{center}
\begin{tabular}{|p{.35\columnwidth}||p{.3\columnwidth}||p{.1\columnwidth}|}
\hline
{DOF} & {Range / Resolution} & \textbf{$R_i$}\\
  \hline \hline
    Oarsmen RX (208) & $0^o$ to $160^o$ / by $1^o$ & 160 \\
  \hline
    Oarsmen RY (208) & $0^o$ to $160^o$ / by $1^o$ & 160 \\
  \hline
      Oarsmen water (208) & on/off & 2 \\
  \hline
Shooters (1,175) & on/off & 2\\
  \hline
lights (6,200) & off or one of 12 colors & 13 \\
  \hline
\end{tabular}
\end{center}
\caption{Estimated fountain system degrees of freedom.} \label{bellagio}
\vspace{-.2in}
\end{table}

Table \ref{bellagio} articulates a model for this system.  For the Oarsman, which rotate about two axes, we assume a range of motion of $160^o$ with a resolution of $1^o$ in each dimension.  We assume the water shooting out of the cannon to be on or off with a single pressure setting.  Likewise, the Shooters, are either on or off without articulation.  The lights can be `off' or one of twelve colors (as modeled by a moderate segmentation of the color wheel).  We ignore the music that plays alongside.

%\textcolor{red}{Figure that shows how adding an extra dof changes expressivity; and how it shows limits of measure to take into account dynamics}\\

Thus, to compute the kinematic mechanization capacity, we find the following computation.
\vspace{-.05in}
\begin{eqnarray}
\mathcal{K}=\log_2(2^{1175+208}\times160^{208+208}\times13^{6200})
=\log_2(4.9\times10^{8239}\mbox{ configurations})
\approx27,372 \mbox{ bits}
\end{eqnarray}

We could argue over which is more interesting to watch: a NAO or the Bellagio fountains, but this metric provides a quantitative bound on \textit{how much more expressive} the fountains are.  In this case, about two orders of magnitude with respect to the amount of information they can encode.  This result is consistent with expressivity scaling with system expense and tourist attendance. %This might strike roboticists as odd, but in terms of system expense and tourist attendance, the measure is consistent.

What if all the water cannons were the articulated, Oarsman variety?   In that case, we have:
\vspace{-.05in}
\begin{eqnarray}
\mathcal{K}=\log_2(2^{1383}\times160^{1383}\times13^{6200})
=\log_2(1.2\times10^{10371}\mbox{ configurations})
\approx34,452 \mbox{ bits}
\end{eqnarray}

\noindent Thus, by upgrading 1,175 cannons to the articulated variety, we don't gain much in expressivity.  If, in addition, we boost the resolution of each cannon of the original system to $0.1^o$, we have:
\vspace{-.05in}
\begin{eqnarray}
\mathcal{K}=\log_2(2^{1383}\times1600^{1383}\times13^{6200})
=\log_2(1.2\times10^{11754}\mbox{ configurations})
\approx39,046\mbox{ bits}.
\end{eqnarray}

\noindent Thus here, we can see how adding water cannons and articulation resolution increases the expressivity of the platform, but we do not capture the additional expressivity that the dynamics of timing and water add to (and take away from) the system.  For example, by moving variable speeds, these fountains can create different patterns in the water, which add to the system's expressivity.  On the other hand, in the presence of water not all points in the cannon's range might be physically feasible.  
\end{document}